\documentclass[sigconf]{acmart}
\usepackage{pdfpages}
\pdfoutput=1
\usepackage{booktabs} 
\usepackage{color}
\usepackage{makecell}
\usepackage{pgfplots}
\usetikzlibrary{matrix}
\usepgfplotslibrary{groupplots}
\pgfplotsset{compat=newest}
\usepackage{tabularx}

\copyrightyear{2019}
\acmYear{2019} 
\setcopyright{iw3c2w3}
\acmConference[WWW '19]{Proceedings of the 2019 World Wide Web Conference}{May 13--17, 2019}{San Francisco, CA, USA}
\acmBooktitle{Proceedings of the 2019 World Wide Web Conference (WWW '19), May 13--17, 2019, San Francisco, CA, USA}
\acmPrice{}
\acmDOI{10.1145/3308558.3313595}
\acmISBN{978-1-4503-6674-8/19/05}

\newcommand{{\modelname}}{\textit{GVNR}}
\newcommand{{\modelnametext}}{\textit{GVNR-t}}

\usepackage{balance}

\begin{document}

\title{Global Vectors for Node Representations}

\author{Robin Brochier}
\authornote{These authors contributed equally to the work.}
\affiliation{%
  \institution{Universit\'e de Lyon, Lyon 2, ERIC EA3083 \\ Digital Scientific Research Technology}
}
\email{robin.brochier@univ-lyon2.fr}

\author{Adrien Guille}
\authornotemark[1]
\affiliation{%
  \institution{Universit\'e de Lyon, Lyon 2, ERIC EA3083}
}
\email{adrien.guille@univ-lyon2.fr}

\author{Julien Velcin}
\affiliation{%
  \institution{Universit\'e de Lyon, Lyon 2, ERIC EA3083 }
}
\email{julien.velcin@univ-lyon2.fr}

\renewcommand{\shortauthors}{Brochier, Guille and Velcin}

\begin{abstract}
Most network embedding algorithms consist in measuring co-occur-rences of nodes via random walks then learning the embeddings using Skip-Gram with Negative Sampling. While it has proven to be a relevant choice, there are alternatives, such as GloVe, which has not been investigated yet for network embedding. Even though SGNS better handles non co-occurrence than GloVe, it has a worse time-complexity. In this paper, we propose a matrix factorization approach for network embedding, inspired by GloVe, that better handles non co-occurrence with a competitive time-complexity. We also show how to extend this model to deal with networks where nodes are documents, by simultaneously learning word, node and document representations. Quantitative evaluations show that our model achieves state-of-the-art performance, while not being so sensitive to the choice of hyper-parameters. Qualitatively speaking, we show how our model helps exploring a network of documents by generating complementary network-oriented and content-oriented keywords. 
\end{abstract}

%
%



\maketitle

\section{Introduction}

Networks are ubiquitous. The Web is a large-scale network of resources, social media help developing broad online social networks \cite{guille2013osn}, the scientific literature forms a vast network of documents, from which one can derive a network of co-authors \cite{tang2008arnetminer}, \textit{etc}. Understanding and exploring these networks involves solving tasks like node classification or link prediction. Efficiently solving these tasks via machine learning requires meaningful representations of the nodes.

The usual approach is to learn node representations using techniques originally devised for word embedding, based on the distributional hypothesis \cite{sahlgren2008distributional}. The analogy between word embedding and network embedding makes sense, because of the similarities in some of the statistical properties of networks and language \cite{perozzi2014deepwalk}. DeepWalk \cite{perozzi2014deepwalk}, arguably the most popular network embedding algorithm, consists in extracting sequences of nodes, akin to sentences, via truncated random walks and then learning node representations based on the Skip-Gram model \cite{mikolov2013distributed}, using the hierarchical softmax approximation. Node2Vec \cite{grover2016node2vec} builds on DeepWalk and suggests another strategy for extracting node sequences via biased truncated random walks. It learns the node representations based on the Skip-Gram model, using the negative sampling approximation (SGNS).
Metapath2vec \cite{dong2017metapath2vec} adapts DeepWalk to heterogeneous networks. Other variants based on SGNS are generalized into a common frame in \cite{qiu2018network}.

Dealing with large-scale networks requires low-complexity algorithms. An issue with these algorithms is that SGNS scales linearly in the size of the corpus of node sequences. GloVe \cite{pennington2014glove}, the main alternative of SGNS, hasn't been investigated yet for network embedding, even though it scales sub-linearly in the size of the corpus. Still, GloVe is limited in the sense that it ignores non co-occurrence, as opposed to SGNS, which could result in less relevant node representations.

In this paper, we address these two issues and propose a matrix factorization approach for network embedding, inspired by GloVe. Our contributions are the following:

\begin{itemize}
\item we present a general model for network embedding, that consists in factorizing a thresholded co-occurrence matrix. By formulating a regression problem on all positive entries and randomly sampled zero entries, this model can take both co-occurrence and non co-occurrence into account, while preserving a competitive time complexity;
\item we show how to extend this general model to networks of documents, by jointly learning word, document and node representations;
\item we quantitatively assess the performance of the general model on well-known networks, against recent baselines. Not only we show that it outperforms GloVe by a very large margin on several datasets, but we also show that it outperforms on-par with recent network embedding algorithms;
\item we quantitatively and qualitatively show that our model brings interesting innovation to deal with network of short documents. We show how to leverage the extension of our model to explore such networks by suggesting keywords.
\end{itemize}

The rest of the paper is organized as follows. In Section~\ref{sec:related_work} we survey related work. We present in details our general model, discuss its relationship to other models, and show how to deal with networks of short documents by incorporating text into the model in Section \ref{sec:model}. Next, in Section \ref{sec:experiments}, we present a thorough experimental study, where we assess the performance of our model following the usual evaluation protocol on a node classification task for well-known networks. We also discuss hyper-parameter sensitivity, and evaluate, quantitatively speaking, the extended model for networks of documents. In Section \ref{sec:case_study}, we present a case study that illustrates the recommendation of keywords with the extended model. Lastly, we conclude this paper and provide future directions in Section~\ref{sec:conclusion}. 

The code for both our model and the evaluation procedure are made publicly available\footnote{\url{https://github.com/brochier/gvnr}}.

\section{Related work}
\label{sec:related_work}

The quality and informativeness of data representation greatly influence the performance of machine learning algorithms. For this reason, a lot of efforts are devoted to devising new ways of learning representations \cite{bengio2013representation}. Word embedding, \textit{i.e.}, the task of learning representations of words, is tightly connected to the task of learning representations of nodes, \textit{i.e.} network embedding. In this section, we first cover important works related to word embedding and then survey recent developments in network embedding.

\subsection{Word Embedding}

The distributional hypothesis is the basis for word embedding. It states that distributional similarity and meaning similarity are correlated, which allows learning representation of words based on the contexts in which they occur, the context of a word being co-occurring words \cite{sahlgren2008distributional}. Co-occurrences are observed by sliding a window over a large corpus.

The Skip-Gram \cite{mikolov2013distributed} model learns word representations by maximizing the log-likelihood of a multiset of co-occurring word pairs, $C$:
\begin{equation} 
\sum_{(w_i,w_j) \in C} \log p(w_j|w_i).
\end{equation}
The conditional probability is given by the softmax function, parameterized by the word vectors:
\begin{equation} 
p(w_j|w_i) = \frac{e^{u_i \cdot v_j}}{\sum_{W} e^{u_i \cdot v_w}}.
\end{equation}
 This formulation is impractical because of the cost of computing the denominator. For this reason, two variants are introduced in \cite{mikolov2013distributed}.
 Skip-Gram with Hierarchical Softmax (SGHS) uses a binary tree to approximate $p(w_j|w_i)$ and speed-up learning. 
 Skip-gram with Negative Sampling (SGNS) redefines $p(w_j|w_i)$ to make it easier to compute:
\begin{equation}
p(w_j|w_i) = \frac{1}{1+e^{-u_i \cdot v_j}} = \sigma(u_j \cdot v_j).
\end{equation}
 The objective becomes maximizing the following log-likelihood:
\begin{equation}
\sum_{(w_i,w_j) \in C} \Big( \log \sigma(w_j \cdot w_i) + \sum_{k=1}^K \mathbb{E}_{w_{k} \sim q(w_{k})} \big [ \log \sigma(-w_k \cdot w_i) \big ] \Big).
\end{equation}
It boils down to a classification task that consists in distinguishing the pairs of co-occurring words in $C$ from random pairs of words, \textit{i.e.} the negative samples. For each $(w_i, w_j) \in C$, $K$ negative samples $(w_i, w_k)$ are drawn, with $q(w_{k}) \propto \text{frequency}(w_k)^{\frac{3}{4}}$.

The GloVe \cite{pennington2014glove} model learns word representations by factorizing the word-word co-occurrence matrix. Its objective is minimizing the reconstruction error, only for positive entries of $X$:
\begin{equation}
\sum_{i=1}^n \sum_{j=1}^n f(x_{ij})\big(u_i \cdot v_j + b^U_i + b^V_j - \log (x_{ij})\big)^2,
\end{equation}
where $f(x_{ij})$ is the following weighting function, that notably reduces the importance of rare co-occurrences and filter-out zero entries:
\begin{equation}
f(x_{ij}) = \begin{cases}
    		\big(x/x_{\text{max}}\big)^{\frac{3}{4}} &\text{~if~} x < x_{\text{max}},\\
    		1 &\text{~otherwise}.
    	\end{cases}
\end{equation}
The authors show that the distribution of $x_{ij}$ follows a power-law and that the time complexity of GloVe is $\mathcal{O}(|C|^{\frac{1}{\alpha}})$, $\alpha$ being the exponent of the power-law. Because $\alpha$ is usually larger than 1 for text, $\frac{1}{\alpha}$ becomes smaller than 1. Hence, this model has a better time complexity than Skip-gram with Negative Sampling, which runs in $\mathcal{O}(|C|)$.

In \cite{levy2015improving}, Levy and Goldberg note that the introduction of a bias for each target/context word adds an extra degree of freedom to the GloVe model as compared to the Skip-gram model.

\subsection{Network Embedding}

Even though the distributional hypothesis originated in linguistics and is naturally leveraged for word embedding, Perozzi \textit{et al.} establish the connection with network embedding. To do so, they show that the frequency at which nodes appear in short random walks follows a power-law distribution, like the frequency of words in language \cite{perozzi2014deepwalk}. 

They propose DeepWalk, that consists in applying skip-gram with hierarchical softmax on a corpus of node sequences, deemed equivalent to sentences, generated with truncated random walks \cite{perozzi2014deepwalk}. For some specific tasks, the representations learned with DeepWalk offer large performance improvements. Thus, many subsequent works focus on modifying or extending DeepWalk. 
Node2vec replaces random walks with biased random walks, in order to better balance the exploration-exploitation trade-off, arguing that the added flexibility in exploring neighborhoods helps learning richer representations \cite{grover2016node2vec}.
Dong \textit{et al.} address the heterogeneous network representation learning problem. They propose Metapath2vec \cite{dong2017metapath2vec}, a modification of DeepWalk based on meta-path-based random walks to generate sequences of heterogeneous nodes. They also suggest learning the representation with negative sampling instead of hierarchical softmax.
In \cite{yang2015network}, Yang \textit{et al.} prove that skip-gram with hierarchical softmax can be equivalently formulated as a matrix factorization problem. They then propose Text-Associated DeepWalk (TADW), to deal with networks of documents. TADW consist in constraining the factorization problem, with a pre-computed representation of documents via LSA \cite{deerwester1990ilsa}.

Qiu \textit{et al.} \cite{qiu2018network} provide the theoretical connections between Skip-Gram based network embedding algorithms and the theory of graph Laplacian. This allows them to unify DeepWalk, LINE \cite{tang2015line} (which they prove to be a special case of DeepWalk), PTE and Node2Vec, all with with negative sampling, into the matrix factorization framework.

\section{Model formulation}
\label{sec:model}

In this section, we present our model, {\modelname} (Global Vectors for Node Representation), to learn node representations, taking into account both co-occurrence and non co-occurrence. We formulate a factorization problem on the thresholded co-occurrence matrix. More specifically, we formulate this problem so that the reconstruction error is measured on all the positive entries and some randomly sampled zero entries. We begin by listing the set of notations we use in Table \ref{tab:notations}, next we describe the matrix to factorize and then formulate the factorization problem. Eventually, we discuss the relationship to other models.

\begin{table}[]
\caption{Notations.}
\begin{tabular}{l|l}
Notation & Definition\\
\hline
$n$ & Number of nodes.\\
$d$ & Embedding dimension.\\
$C$ & Corpus of co-occurring nodes.\\
$U \in \mathbb{R}^{n \times d}$ & Target node embeddings. \\
$V \in \mathbb{R}^{n \times d}$ & Context node embeddings. \\
$l \in \mathbb{N}^+_*$ & Window size for observing co-occurrence. \\
$X \in \mathbb{R}^{n \times n}$ & Co-occurrence matrix. \\
$n_i$ & Number of nodes that co-occur with node $i$. \\
\end{tabular}
\label{tab:notations}
\end{table}

\subsection{Description of the Matrix to Factorize}

We observed node co-occurrences in truncated random walks \cite{perozzi2014deepwalk}. Then, for each node $j$ visited within $q$ steps, with $q \leq l$, from a node $i$, we increase $X_{ij}$ by $\frac{1}{q}$ \cite{pennington2014glove}. Thus, we construct a weighted co-occurrence matrix, so that distant co-occurrences are increasingly downweighted. The matrix obtained with this procedure is approximately proportional to the weighted sum of the $l$ first of the adjacency matrix: $\sum_{i=1}^l \frac{1}{i} A^i$. However, this sum is likely to give a denser co-occurrence matrix because unlikely and distant co-occurrences will lead to coefficients close to zero. On the contrary, the truncated random walk is likely to estimate these coefficients as exactly zero. Because we consider coefficients close to zero as noise, we're not interested in calculating them, and can computationally benefit from a sparser matrix. For the same reason, we zero-out coefficients that are less than a threshold $x_{\min}$, assuming they are irrelevant.

\subsection{Formulation of the Factorization Problem}

We formulate a factorization problem on $X$, measuring the error only for positive coefficients and a fraction of randomly sampled zero coefficients. Note that we measure the error w.r.t the logarithm, which help compress the range of values in $X$ \cite{pennington2014glove}. Because the matrix is already thresholded, we assume all the remaining positive entries have the same importance, thus we don't weight the least-square objective:
\begin{equation}
\underset{U,V,b^U,b^V}{\mathrm{argmin}} \sum_{i=1}^n \sum_{j=1}^n s(x_{ij})\big(u_i \cdot v_j + b^U_i + b^V_j - \log (c + x_{ij})\big)^2.
\end{equation}
The constant $c \in ]0;1]$ allows for smoothing $X$ while making the logarithm negative when $x_{ij} = 0$. The function $s$ effectively selects the coefficients considered for measuring the reconstruction error:
\begin{equation}
s(x_{ij}) = \begin{cases}
    		1 & \text{if } x_{ij} > 0,\\
    		m_{i} & \text{else, with } m_{i} \sim \text{Bernoulli}(\alpha_i).
    	\end{cases}
\end{equation}
It takes the value 1 for all positive coefficients of $X$, while for zero coefficients, its value is given by a Bernoulli random variable, $m_i$, with a node-specific parameter $\alpha_i$. Denoting the proportion of positive coefficient on the i$^{\text{th}}$ row of $X$ by $p_i$, $\alpha_i$ is calculated in terms of the odd-ratio:
\begin{equation}
\alpha_i = \begin{cases}
    		k \times \frac{p_i}{1-p_i} & \text{if } p_i \leq (k+1)^{-1},\\
    		1 & \text{else}
    	\end{cases}
\end{equation}
where $k > 0$ is an hyper-parameter that controls the proportion of zero coefficients incorporated into the calculation of the reconstruction error, akin to the number of negative samples in SGNS. The larger $k$, the more importance is given to pushing away vectors of non co-occurring nodes. 

\subsection{Relationship to Other Models}

This objective function bears a resemblance to the objective of GloVe, still the two are quite different. As stated by the equation 5, GloVe performs a weighted least-square regression on the raw co-occurrence matrix. Although the authors of GloVe claim that rare co-occurrences are noisy and carry less information than the more frequent ones, this objective function still takes them into account (with a proportionally smaller weight, according to the equation 6). Our model has a stronger interpretation of this claim, by zeroing out rare co-occurrences, and weighting equally all the others. In addition, while all zero coefficients are ignored in GloVe due to the definition of $f$ in equation 6, we incorporate a fraction of them, proportional to the quantity of positive coefficients. That constitutes an additional set of constraints we think should lead to better representations. Lastly, thresholding $X$ helps us eliminating noise while drastically sparsifying it, since $x_{ij}$ follows a power law \cite{perozzi2014deepwalk, pennington2014glove}.

The complexity of Skip-Gram based algorithms, implemented with the procedure described in \cite{mikolov2013distributed}, is linear in the size of the multiset of pairs of co-occurring nodes, \textit{i.e.} $\mathcal{O}(|C|)$. Based on the proof given in \cite{pennington2014glove}, the complexity of our model is $ \mathcal{O}(|C|^{\frac{1}{\alpha}})$, where $\alpha$ is the exponent of the power law that models $x_{ij}$. For the networks studied in this paper, we observe a mean value for $\frac{1}{\alpha}$ of $0.79$.

\subsection{Extension to Networks of Documents}

Lastly, we show how to extend the general model under the name {\modelnametext}, to deal with networks where nodes are text documents.

Assuming word order is negligible for documents \cite{deerwester1990ilsa}, we can model a text as a bag of words and thus represent it by a vector $\delta \in \mathbb{N^+}^m$, $m$ being the size of the vocabulary. We can further assume that the meaning of a text can be captured by averaging the representations of its words \cite{le2014paragraph}. Therefore, with $W \in \mathbb{R}^{m \times d}$ a word embedding matrix, rather than learning the the context-vector of a node as explained previously, we define it as the average of the representations of the words it contains:

\begin{equation}
\underset{U,V,b^U,b^V}{\mathrm{argmin}} \sum_{i=1}^n \sum_{j=1}^n s(x_{ij})\big(u_i \cdot \frac{\delta_j ~ W}{|\delta_j|_1} + b^U_i + b^V_j - \log (c + x_{ij})\big)^2,
\end{equation}
where $\delta_j$ is the bag of words representation of text associated to the node $j$, and $|\delta_j|_1$ is the number of words in it. Thus, the model jointly learns node and word representations, that in turn allows representing documents.

\section{Quantitative evaluation}
\label{sec:experiments}

A common task in network analysis is node classification. Following the experimental designs in recent works \cite{yang2015network, qiu2018network}, we assess the quality of the representations learned with {\modelname} by using them as input of a linear classifier to solve multi-class and multi-label classification tasks.

\subsection{Networks}    

\begin{table}[]
\small
\caption{General properties of the studied networks.}
\begin{tabular}{l|ccccc}
 &  $|V|$  & $|E|$ & \# labels & weighted & multi-label \\ \hline
Citation 1 & 2,708 & 10,556 &7 & no &  no \\
Citation 2 & 3,312 & 9,226 & 6 & no & no \\
Co-authorship & 5,021  & 29,856 & 5 & yes &  no \\
Protein & 3,890  & 76,584  & 50 & no & yes\\
Language & 4,777  & 184,812 & 40  & yes  & yes
\end{tabular}
\label{tab:networks}
\end{table}

To show the versatility of {\modelname}, we consider five networks of various nature:
\begin{itemize}
    \item Two citations networks: Citation (1) and Citation (2), extracted respectively from Cora and Citeseer\footnote{https://linqs.soe.ucsc.edu/data}; each node is an article and is labelled with a conference. 
    \item A co-authorship network extracted from DBLP\footnote{https://dblp.uni-trier.de/}; each node is an author labelled with a domain of expertise\footnote{https://static.aminer.org/lab-datasets/expertfinding/} and each edge is weighted according to the number of common publications. 
    \item A protein-protein interaction (PPI) network \cite{stark2010biogrid} which is a subgraph of the PPI network for Homo Sapiens. Each node is associated to several labels that represent biological states. 
    \item A language network that describes collocated words observed in the first $10^8$ bytes of the English Wikipedia\footnote{http://mattmahoney.net/dc/text.html} (as of March, 2006); each node is associated to multiple labels, which are the potential part-of-speech tags identified with the Stanford POS tagger \cite{toutanova2003feature}.
\end{itemize}

The general properties of these five networks are reported in Table \ref{tab:networks}.

\subsection{Tasks and Evaluation Metrics}

For each network, we consider a classification task and evaluate the performance of a linear classifier, namely a logistic regression, using node representations as input. We use the LIBLINEAR implementation \cite{fan2008liblinear}, without regularization, for a fair comparison between different representations. For each network, we train and evaluate the classifier by cross-validation, with varying split ratios.

More precisely, we consider a multi-class classification problem for the citation and co-authorship networks and measure the overall accuracy of the one-vs-all logistic regression. We consider a multi-label classification problem for the protein-protein and the language networks and measure the $F_1$ score of the one-vs-all logistic regression, following the procedure described in \cite{qiu2018network}.

\subsection{Compared Representations}

We run $\gamma=80$ random walks per node of length $t=40$ and apply a sliding window of size $l=5$ to generate a multiset of co-occurring nodes and then learn representation with $d=80$ with four algorithms:

\begin{itemize}
    \item DeepWalk: we report the results with the hierarchical softmax approximation.
    \item GloVe: we report the results with $x_{\max} = 10$.
    \item NetMF: we report the results with the small-scale implementation provided by the authors, with $k=10$ negative samples.
    \item {\modelname}: we report the results with $c=1$, $x_{\text{min}}=1$ and $k=1$ and discuss their impact in the next sub-section.
\end{itemize}

\subsection{Result Analysis}

Tables \ref{table:citation1} to \ref{table:protein} detail the accuracy measures.
The classifier performs well with the representations learned by {\modelname}, achieving similar or better results w.r.t the representations learned with DeepWalk and NetMF. Still, there is one exception concerning the language network, where the classifier performs best with the representation learned with GloVe. We sum up our findings in two points:

\begin{enumerate}
\item {\modelname} always achieves a good performance, while there is more variance in the performance of the baselines. For instance, NetMF is largely outperformed by the others on the language network, GloVe is also significantly outperformed on the co-authorship network, while DeepWalk struggles to learn good representations from the citation (2) network.

\item Thresholding always improves the performance of {\modelname}. As an example, its leads to an average gain of 9.4 accuracy on the co-authorship network.
\end{enumerate}

Interestingly, GloVe produces the best results on the language network. The average number of neighbors is $38.7$ for this network which makes it the denser network.

\begin{table}[h]
\center
\caption{Accuracy on the citation (1) network.}
\label{table:citation1}
\begin{tabular}{l|ccccc}
          & \multicolumn{5}{c}{\% of training data}\\
 & 10\%    &20\%    & 30\%    & 40\%    & 50\%   \\ \hline
GloVe &57.7   &62.4   &69.5   &72.8   &73.8  \\ 
{\modelname} ($x_{\text{min}}=0$) &58.5   &62.5   &70.7   &73.4   &75.0  \\
NetMF &65.7   &\textbf{72.9}   &\textbf{76.4}   &\textbf{78.6}   &79.4   \\ 
DeepWalk &67.8   &71.6   &74.5   &75.8   &79.2 \\
{\modelname} ($x_{\text{min}}=1$) &\textbf{69.5}   &72.6   &75.9   &78.1   &\textbf{80.2} 
\end{tabular}
\end{table}

\begin{table}[h]
\center
\caption{Accuracy on the citation (2) network.}
\label{table:citation2}
\begin{tabular}{l|ccccc}
          & \multicolumn{5}{c}{\% of training data}\\
 & 10\%    &20\%    & 30\%    & 40\%    & 50\%   \\ \hline
GloVe &42.8   &53.5   &55.3   &56.2   &56.8 \\ 
{\modelname} ($x_{\text{min}}=0$)&38.7   &46.8   &49.1   &50.4   &50.9 \\
NetMF &\textbf{51.2}   &54.8   &55.1   &55.0   &54.8\\ 
DeepWalk &41.3   &52.5   &54.5   &55.5   &56.0\\
{\modelname} ($x_{\text{min}}=1$) &45.6   &\textbf{55.6}   &\textbf{57.3}   &\textbf{58.7}   &\textbf{59.0}
\end{tabular}
\end{table}

\begin{table}[h]
\center
\caption{Accuracy on the co-authorship network.}
\label{table:coautorship}
\begin{tabular}{l|ccccc}
          & \multicolumn{5}{c}{\% of training data}\\
 & 10\%    &20\%    & 30\%    & 40\%    & 50\%   \\ \hline
GloVe &41.0   &42.1   &43.7   &46.4   &51.2 \\ 
{\modelname} ($x_{\text{min}}=0$)  &60.3   &64.6   &67.4   &67.1   &68.2\\
NetMF  &60.7   &66.2   &70.1   &72.1   &72.8 \\ 
DeepWalk &\textbf{75.4}   &\textbf{77.2}   &\textbf{77.3}   &\textbf{75.9}   &\textbf{79.3} \\
{\modelname} ($x_{\text{min}}=1$)  &74.7   &75.3   &76.3   &73.8   &74.6 \\ 
\end{tabular}
\end{table}

\begin{table}[h]
\center
\caption{$F_1$ score on the language network.}
\label{table:language}
\begin{tabular}{l|ccccc}
          & \multicolumn{5}{c}{\% of training data}\\
 & 10\%    &20\%    & 30\%    & 40\%    & 50\%   \\ \hline
GloVe &\textbf{34.0}   &\textbf{44.1}   &\textbf{46.7}   &\textbf{47.7}   &\textbf{48.6} \\ 
{\modelname} ($x_{\text{min}}=0$) &31.7   &40.7   &43.2   &44.7   &45.1 \\
NetMF  &27.5   &33.5   &36.2   &37.7   &38.7 \\ 
DeepWalk &33.6   &43.6   &46.2   &47.6   &48.2\\
{\modelname} ($x_{\text{min}}=1$) &32.2   &41.7   &44.0   &45.2   &46.1 \\ 
\end{tabular}
\end{table}

\begin{table}[h]
\center
\caption{$F_1$ score on the protein-protein network.}
\label{table:protein}
\begin{tabular}{l|ccccc}
          & \multicolumn{5}{c}{\% of training data}\\
 & 10\%    &20\%    & 30\%    & 40\%    & 50\%   \\ \hline
GloVe &11.8   &13.8   &15.6   &17.3   &19.1  \\ 
{\modelname} ($x_{\text{min}}=0$) &10.7   &13.3   &15.0   &16.7   &18.1 \\
NetMF   &10.2   &11.7   &13.5   &14.7   &16.1 \\ 
DeepWalk &\textbf{12.2}   &\textbf{13.9}   &\textbf{16.2}   &\textbf{17.8}   &\textbf{19.6} \\
{\modelname} ($x_{\text{min}}=1$) &11.7   &13.3   &15.8   &17.6   &19.2 \\ 
\end{tabular}
\end{table}

\subsection{Impact of the Hyper-Parameters}

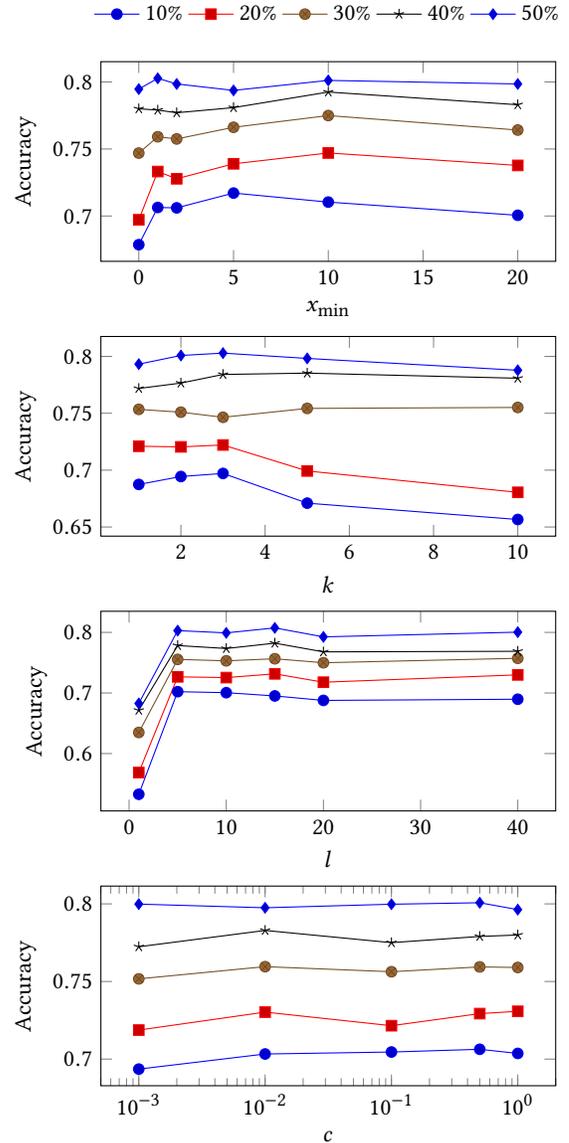
\begin{figure}[h]
\center
\begin{tikzpicture}
\begin{groupplot}[group style={group size=1 by 4},height=0.5\columnwidth,width=0.9\columnwidth]
    \nextgroupplot[ylabel={Accuracy}, xlabel={$x_{\text{min}}$}, legend style={legend columns=-1, draw=none} ,legend to name=main]
    \foreach \column in {1,...,5}{\addplot+[]table[col sep=comma,x={filters},y={0.\column}]{xmin.csv};}
    \addlegendentry{10\%};\addlegendentry{20\%};\addlegendentry{30\%};\addlegendentry{40\%};\addlegendentry{50\%};
    \coordinate (top) at (rel axis cs:0,1);
    \nextgroupplot[ylabel={Accuracy}, xlabel={$k$}]
    \foreach \column in {1,...,5}{\addplot+[]table[col sep=comma,x={filters},y={0.\column}]{k.csv};}
    \nextgroupplot[ylabel={Accuracy}, xlabel={$l$}]
    \foreach \column in {1,...,5}{\addplot+[]table[col sep=comma,x={filters},y={0.\column}]{l.csv};}
    \nextgroupplot[ylabel={Accuracy}, xlabel={$c$}, xmode=log]
    \foreach \column in {1,...,5}{\addplot+[]table[col sep=comma,x={filters},y={0.\column}]{c.csv};}
    \coordinate (bot) at (rel axis cs:1,0);
  \end{groupplot}
\node[above=2em, right=-8.25em, inner sep=0pt] at(top -| current bounding box.north) {\pgfplotslegendfromname{main}};
\end{tikzpicture}
\caption{Sensitivity of {\modelname} to the hyper-parameters $x_{\text{min}}$, $k$, $l$ and $c$ on the citation (1) network.}
\label{figure:sensitivity}
\end{figure}

Figure \ref{figure:sensitivity} shows the sensitivity, in terms of accuracy, of {\modelname} to its hyper-parameters, namely, the threshold $x_{\min}$, the shifting constant $c$, the window length $l$, and the sampling proportion $k$. We only report the results for the citation (1) network since their are similar across the considered networks. We see that $c$ has little impact on the accuracy. In practice, we found {\modelname} performs best across the datasets with $c=1$. It seems that the model performs best with a threshold value between 1 and 5. Setting $x_{\min} > 0$ clearly improves the performance. In general, we observed that $k=1$ constantly brings an improvements over $k=0$, but higher values only brings a boost for high training ratios. Finally, the accuracy increases along with the window size $l$, with few improvements above $l=5$.  

\subsection{Additional Results with Text}

We now report additional results when taking into account the text information, associated to the nodes of the citation networks, \textit{i.e.} titles and abstracts. We consider three ways of learning text-aware representations: TADW \cite{yang2015network}, the representations learned with DeepWalk concatenated with the LSA representation of the texts, and {\modelnametext}. We report the results for TADW with $20$ iterations and 4 iterations for {\modelnametext}.

Tables \ref{table:citation1-text} and \ref{table:citation2-text} report the accuracies. {\modelnametext} shows interesting improvements over the simple concatenation of text and graph features and the reference baseline in the field, TADW. This motivates the study of the word representations learned by {\modelnametext} and the interplay between the node and document representations.

\begin{table}[h]
\center
\caption{Accuracy on the citation (1) network, considering the text features.}
\label{table:citation1-text}
\begin{tabular}{l|ccccc}
          & \multicolumn{5}{c}{\% of training data}\\
 & 10\%    &20\%    & 30\%    & 40\%    & 50\%   \\ \hline
LSA &54.7   &61.0   &62.4   &63.0   &62.8 \\ 
DeepWalk+LSA &73.8   &77.9   &78.4   &78.1   &78.1 \\
TADW &77.1   &78.8   &78.2   &78.8   &78.6 \\
{\modelnametext} &\textbf{79.3}   &\textbf{80.7}   &\textbf{80.8}   &\textbf{81.4}   &\textbf{81.1} \\
\end{tabular}
\end{table}

\begin{table}[h]
\center
\caption{Accuracy on the citation (2) network, considering the text features.}
\label{table:citation2-text}
\begin{tabular}{l|ccccc}
          & \multicolumn{5}{c}{\% of training data}\\
 & 10\%    &20\%    & 30\%    & 40\%    & 50\%   \\ \hline
LSA &52.0   &54.7   &54.7   &58.4   &65.7 \\ 
DeepWalk+LSA &58.3   &60.7   &61.1   &60.0   &61.2 \\
TADW &60.6   &60.1   &60.1   &66.2   &69.3 \\
{\modelnametext} &\textbf{63.3}   &\textbf{62.5}   &\textbf{64.9}   &\textbf{68.6}   &\textbf{70.4}\\
\end{tabular}
\end{table}

\section{Case study}
\label{sec:case_study}

To showcase the usefulness of learning jointly word, node and document representation, we apply {\modelnametext} to  full DBLP network that consists in 1,397,240 documents and 3,021,489 citation relationships. After computing $X$, we threshold it with $x_{\text{min}}=20$, which divides the density of $X_{(ij)}$ by 50. For the learning phase, we keep the same settings as before, that is $d=80$, $k=1$, $c=1$, $l=5$. Note that we apply a standard pre-processing, that consists in merging recurrent phrases, as suggested in \cite{mikolov2013distributed}. Computing $X$ and estimating $U$ and $W$, the node and word representations takes about 8 hours on a single machine with 32 cores and 192 GB of RAM.

Our aim is to show that we can measure the similarity, on the one hand, between word and node representations, and, on the other hand, between word and document representations, to extract sets of complementary keywords.

Tables \ref{table:quali1} and \ref{table:quali2} show 2 randomly selected papers for which we computed the 5 closest word embeddings $w_k$ to, respectively, (i) the node representation $u$ and (ii) its content representation $v$.
First, we note that the keywords are all relevant, even though none of them actually appear in the documents. Then, we see that words close to the node representation are more general, giving a broad view of the studied topic, whereas words close to the document representation are more specialized and provide a more fine-grained perception. We suspect that the network topology, mostly influencing $U$, helps locating a node in its broader context, while the content of documents, mostly influencing $W$, helps in selecting some very specific keywords.
In future work, we would like to investigate further the quality of the learned representations for a wider range of recommendation tasks.

\begin{table}[h]
\center
\footnotesize
\caption{Keyword recommendation by selecting the closest word embeddings $w_k$ to both embeddings $u$ (node) and $v$ (content) of an input document (1).}
\label{table:quali1}
\begin{tabularx}{\columnwidth}{|p{0.15\columnwidth}p{0.76\columnwidth}|}
\hline
Document & \textbf{A brief survey of computational approaches in social computing} Web 2.0 technologies have brought new ways of connecting people in social networks for collaboration in various on-line communities. Social Computing is a novel and emerging computing paradigm... \\
 & \\
Closest words to $u$ (node) & \textit{cold start problem, storylines, document titles, movielens data, computational humor}\\
& \\
Closest words to $v$ (content) & \textit{social, social network, enron email corpus, social networks, extremely large datasets, sites blogs}\\ \hline
\end{tabularx}
\end{table}     

\begin{table}[h]
\center
\footnotesize
\caption{Keyword recommendation by selecting the closest word embeddings $w_k$ to both embeddings $u$ (node) and $v$ (content) of an input document (2).}
\label{table:quali2}
\begin{tabularx}{\columnwidth}{|p{0.15\columnwidth}p{0.76\columnwidth}|}
\hline
Document & \textbf{Discovering company revenue relations from news} A network approach Large volumes of online business news provide an opportunity to explore various aspects of companies. A news story pertaining to a company often cites other companies. Using such company citations we... \\
 & \\
Closest words to $u$ (node) & \textit{datenbanksystemen, denizens, want hear, technological infrastructures, asynchronous discussions} \\
 & \\
Closest words to $v$ (content) & \textit{company, consumer brand, today highly competitive, data cleaning, consumer heterogeneity} \\ \hline
\end{tabularx}
\end{table}        

\section{Conclusion}
\label{sec:conclusion}

In this paper, we presented {\modelname}, a matrix factorization based method for network embedding, that better handles non co-occur-rence than GloVe. We further extended this model to incorporate textual content associated with the nodes to learn meaningful representations of words and documents. We showed that {\modelname} performs state-of-the-art results on a wide range of networks and can provide recommendations based on the distance between words, documents and graph embeddings. In future works we would like to explore further the relations between the word embeddings learned with {\modelnametext} and traditional word embeddings learned with GloVe or Skip-Gram. More particularly, we would like to study whether {\modelnametext} could help learning better word embeddings from small structured corpora.

\bibliographystyle{ACM-Reference-Format}
\balance 
\bibliography{main}

\end{document}